\documentclass[10pt,twocolumn,letterpaper]{article}

\usepackage[pagenumbers]{cvpr} 

\definecolor{cvprblue}{rgb}{0.21,0.49,0.74}
\usepackage[pagebackref,breaklinks,colorlinks,allcolors=cvprblue]{hyperref}
\usepackage{placeins}
\def\paperID{14561} 
\def\confName{CVPR}
\def\confYear{2026}

\title{
DenseAnnotate: Enabling Scalable Dense Caption Collection \\ for Images and 3D Scenes via Spoken Descriptions
}

\author{Xiaoyu Lin \quad Aniket Ghorpade \quad Hansheng Zhu \quad Justin Qiu \quad Dea Rrozhani \quad Monica Lama \\ Mick Yang \quad Zixuan Bian \quad Ruohan Ren \quad Alan B. Hong \quad Jiatao Gu \quad Chris Callison-Burch\\
University of Pennsylvania
}

\begin{document}

\twocolumn[{%
\renewcommand\twocolumn[1][]{#1}%
\maketitle
\begin{center}
  \includegraphics[width=\linewidth]{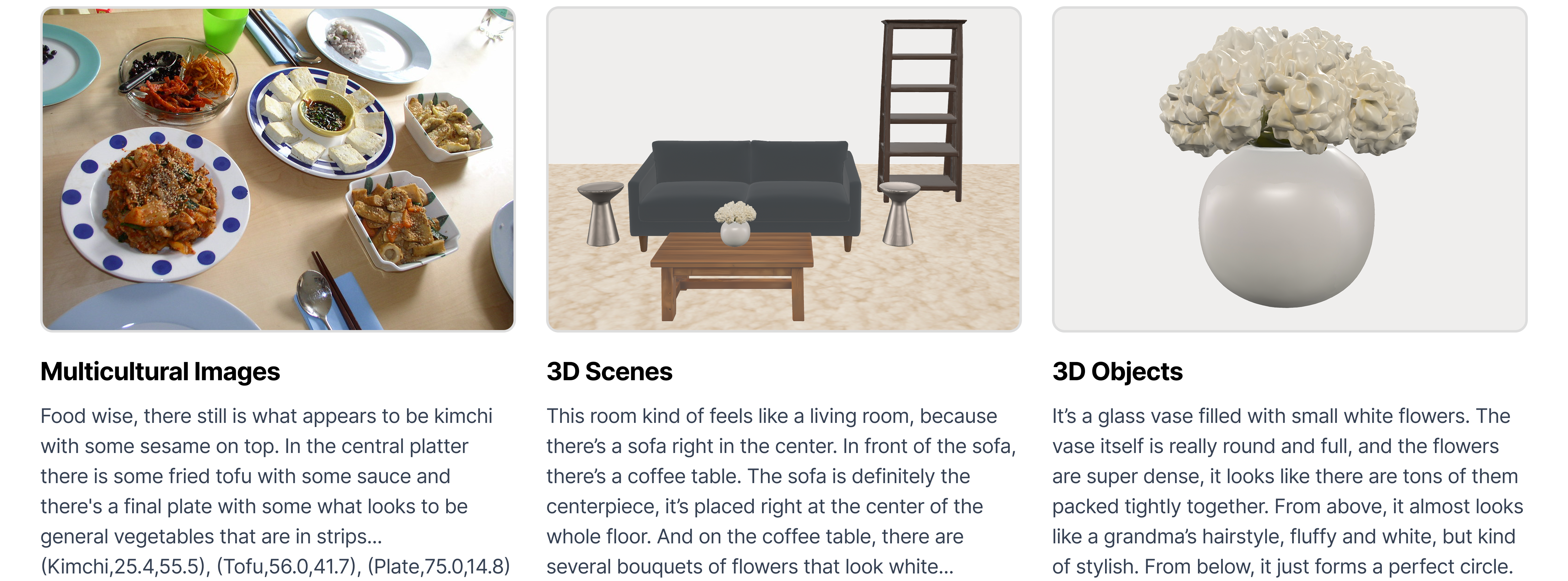}
  
  \captionof{figure}{\textbf{Overview of our Multilingual Dense Captioning (MLDC) dataset.} It includes multicultural images with keypoint annotations aligned with the captions, as well as annotated 3D scenes and their corresponding 3D objects. Partly omitted for clarity.}
  \label{fig:teaser}
\end{center}
}]

\begin{abstract}

With the rapid adoption of multimodal large language models (MLLMs) across diverse applications, there is a pressing need for task-centered, high-quality training data. 
A key limitation of current training datasets is their reliance on sparse annotations mined from the Internet or entered via manual typing that capture only a fraction of an image's visual content. 
Dense annotations, which provide longer and more comprehensive descriptions covering substantially more visual details, attributes, and relationships, are more valuable but remain scarce. 
Traditional text-based annotation pipelines are poorly suited for creating dense annotations: typing limits expressiveness, slows annotation speed, and underrepresents nuanced visual features, especially in specialized areas such as multicultural imagery and 3D asset annotation. 
In this paper, we present \textbf{DenseAnnotate}, an audio-driven online annotation platform that enables efficient creation of dense, fine-grained annotations for images and 3D assets. 
Annotators narrate observations aloud while synchronously linking spoken phrases to image regions or 3D scene parts. 
Our platform incorporates speech-to-text transcription and region-of-attention marking. 
To demonstrate the effectiveness of DenseAnnotate, we conducted case studies involving over \textbf{1,000} annotators across two domains: culturally diverse images and 3D scenes. 
We curate a human-annotated multi-modal dataset of 3,531 images, 898 3D scenes, and 7,460 3D objects, with audio-aligned dense annotations in \textbf{20 languages}, including \textbf{8,746} image captions, \textbf{2,000} scene captions, and \textbf{19,000} object captions. 
Models trained on this dataset exhibit improvements of 5\% in multilingual, 47\% in cultural alignment, and 54\% in 3D spatial capabilities.
Our results show that our platform offers a feasible approach for future vision-language research and can be applied to various tasks and diverse types of data.
\end{abstract}    
\section{Introduction}
\label{sec:intro}

MLLMs have begun to play a role in fields as wide ranging as medicine, robotics, and cultural understanding \citep{kalpelbe2025visionlanguagemodelsmedicine,  li2024visionlanguagefoundationmodelseffective,liu2025culturevlmcharacterizingimprovingcultural}. To be successfully applied to such diverse applications, training data is crucial for MLLMs. Despite the abundance of publicly available data, high-quality training data is still scarce \citep{dong2025scalablevisionlanguagemodel}. The scarcity of high-quality training datasets adds to the difficulty of continuously improving the performance of future MLLMs \citep{li2025surveystateartlarge}. 

Recently, dense captioning has gained increasing attention as an effective strategy to enhance model performance. Dense captions are detailed textual descriptions that capture a richer set of visual elements. For example, PixelProse is a dataset of dense image captions with high quality and fidelity \citep{singla2024pixelsproselargedataset}. Also, increasing caption density has been shown to improve vision-language models' compositional reasoning \citep{doveh2023densealignedcaptionsdac}. Pyramid-XL generates point-language dense captions used to finetune the model, resulting in significantly improved 3D object generation \citep{qi2025gpt4pointunifiedframeworkpointlanguage}. However, these synthetic captioning methods have limitations, such as hallucinations \citep{singla2024pixelsproselargedataset, zhang2025lowhallucinationsyntheticcaptionslargescale}.

Human-annotated data can effectively mitigate the problem. Several studies have published datasets or benchmarks of human-annotated dense captioning images \citep{urbanek2024pictureworth77text}. However, due to limitations in annotation interfaces (\eg reliance on typing and limited interaction modalities), there remains significant room for quality and efficiency improvement. Recent work like COTALK \citep{shen-etal-2025-chain} demonstrates that speech-based annotation significantly improves efficiency, achieving a 40\% speedup over typing-based annotation.

In this work, we publish an online dense captioning toolkit, which has a well-established mechanism to ensure the acquisition of high-quality human-annotated multimodal data. It represents a comprehensive pipeline for multimodal data acquisition, encompassing task design, task deployment, annotation gathering, and releasing the final, high-quality data. \Cref{fig:teaser} illustrates examples from our collected data.

Previous research Molmo and its accompanying PixMo datasets \citep{deitke2024molmopixmoopenweights} mark a significant step forward in developing state-of-the-art open-weight Vision-Language Models (VLMs). Their contribution is particularly exemplified by the adoption of an audio-driven annotation paradigm, which enables the collection of large-scale, high-quality 2D image data without relying on proprietary VLM distillation.
However, PixMo has not released an open-source annotation platform, which limits the community’s ability to expand the dataset in specific domain. Our open-source online platform, DenseAnnotate, distinguishes itself by extending this dense captioning approach into fundamentally new and challenging domains. Crucially, while PixMo focuses on 2D imagery through separate tasks (PixMo-Cap for dense captioning and PixMo-Points for pointing), DenseAnnotate introduces a unified online annotation platform that enables the simultaneous collection of dense captions and pointing annotations within a single interface, and further extends dense captioning capabilities to complex 3D assets. By constructing our MLDC-3D dataset, we directly address the critical lack of scene-level high-quality data required for models dedicated to 3D understanding. Furthermore, DenseAnnotate places a specialized emphasis on capturing cultural nuances through our MLDC-MC dataset. This collection leverages native-language annotators to acquire detailed and culturally aligned descriptions, resulting in significantly longer captions, thus ensuring a depth of linguistic and cultural representation that complements existing open MLLMs efforts.

To summarize, our contributions are three-fold: (1) We develop a unified multimodal annotation platform capable of handling both 2D images and 3D assets in a web-based interface; (2) To the best of our knowledge, we release the first human-annotated multilingual dataset of multicultural 2D imagery with dense captions aligned to pointing annotations, enabling improved multilingual multicultural captioning and region-level grounding; (3) We further construct the first human-annotated multilingual dataset that provides full-scene 3D dense captions together with independent object-level dense descriptions, where each object is physically separable from the scene and semantically grounded within it. This dataset supports 3D understanding and reasoning over point clouds.

\section{Related Work}
\label{sec:related_work}

\begin{figure*}[!t]
  \includegraphics[width=\textwidth]{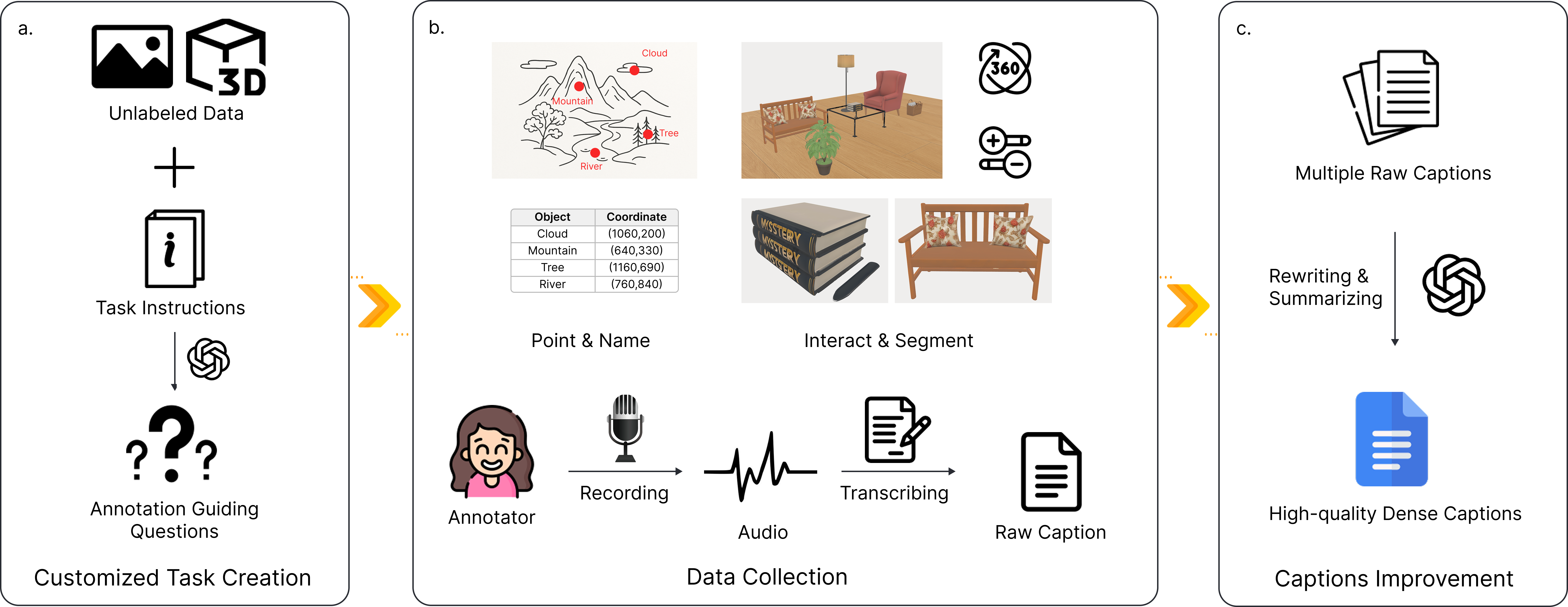}
  \caption{\textbf{Dense captioning generation.} a. Customized Task Creation: After task creators upload unlabeled images or 3D scenes, the platform will generate advice to guide annotators. b. Data Collection: Annotators can record audio while performing Point \& Name on images or Interact \& Segment on 3D scenes. c. Captions Improvement: Summarizing into high-quality captions.}
  \label{fig:spirtfigure}
\end{figure*}

\textbf{Multilingual and Multicultural Image Captioning.}
Most vision-language datasets for image description remain heavily English-centric \citep{nguyen2025multilingualdiversityimprovesvisionlanguage}. Recent years have seen efforts to broaden this scope: for example, the CVLUE benchmark \citep{wang-etal-2024-cvlue} introduces large-scale Chinese vision–language data, enabling systematic study of bilingual capabilities in modern vision–language models. Similarly, the CVQA benchmark \citep{romero2024cvqaculturallydiversemultilingualvisual} provides a culturally diverse multilingual VQA dataset with 10k human-annotated questions from 30 countries covering 31 languages, collected by native speakers to ensure broad cultural coverage. 

However, while these benchmarks are essential for assessment, there remains a critical need for high-quality training data that supports rich, fine-grained descriptions for foundational model training. Our work solved this problem and supports visual grounding by pointing .

\noindent \textbf{3D Scene Captioning.}
Describing 3D environments in natural language is a very recent endeavor, and existing 3D captioning resources are scarce. Recent advances like MMScan \citep{lyu2025mmscanmultimodal3dscene} and TOD3Cap \citep{jin2024tod3cap3ddensecaptioning} utilize semi-automatic pipelines leveraging visual-language models and subsequent human refinement to generate large-scale 3D scene descriptions. However, these approaches are predominantly monolingual (English) and are intrinsically limited by the expressiveness and speed constraints of text-based inputs. In stark contrast, our work introduces an innovative audio-driven annotation platform, enabling the collection of rich, fine-grained, and multilingual captions while ensuring annotation quality by transcription summarization and the platform's auto-check mechanism.

PointLLM \citep{xu2024pointllmempoweringlargelanguage}, a multi-modal large language model designed to understand 3D objects represented as point clouds. Unlike traditional models that rely on 2D images and struggle with issues like depth ambiguity and occlusions, PointLLM directly processes 3D geometric and appearance data. PointLLM is exclusively trained on 3D objects, can handle scene-level point clouds, but it exhibits only limited capability for captioning, let alone complex tasks. The PointLLM paper highlights that effectively handling scene-level point clouds remains constrained by the lack of high-quality annotated data. Precisely addressing this gap, our work provides the essential scene-level high-quality data.

\section{DenseAnnotate}

In this section, we first briefly introduce the dense captioning platform. Then, we show the details of our multilingual dense captioning dataset for multicultural images and 3D assets.

\subsection{DenseAnnotate: A Dense Captioning Toolkit}

\paragraph{Annotation Workflow}
\Cref{fig:spirtfigure} illustrates our system's three-stage workflow. In the first stage, unlabeled images or 3D data are combined with task-specific instructions to generate annotation-guiding questions. Task creators can either adopt the system’s suggested questions or input their own guidance for annotators. In the second stage, human annotators interact with the data through multiple operations—such as pointing and naming for 2D images, and zooming, panning, and freely rotating for 3D objects or scenes. In the latter case, annotators can also isolate and closely examine individual objects within the scene before producing their audio descriptions. To make the captions dense enough, we set the minimum recording times as follows: 60 seconds for images or 3D scenes, and 20 seconds for 3D objects. The recordings are transcribed into raw captions, and annotators can edit typos. We recruit multiple annotators to label each of the images or 3D assets. In the last stage, multiple raw captions are summarized using LLMs to produce high-quality, dense captions, thereby enhancing both the accuracy and richness of the annotations.

\begin{figure*}[!t]
  \includegraphics[width=\textwidth]{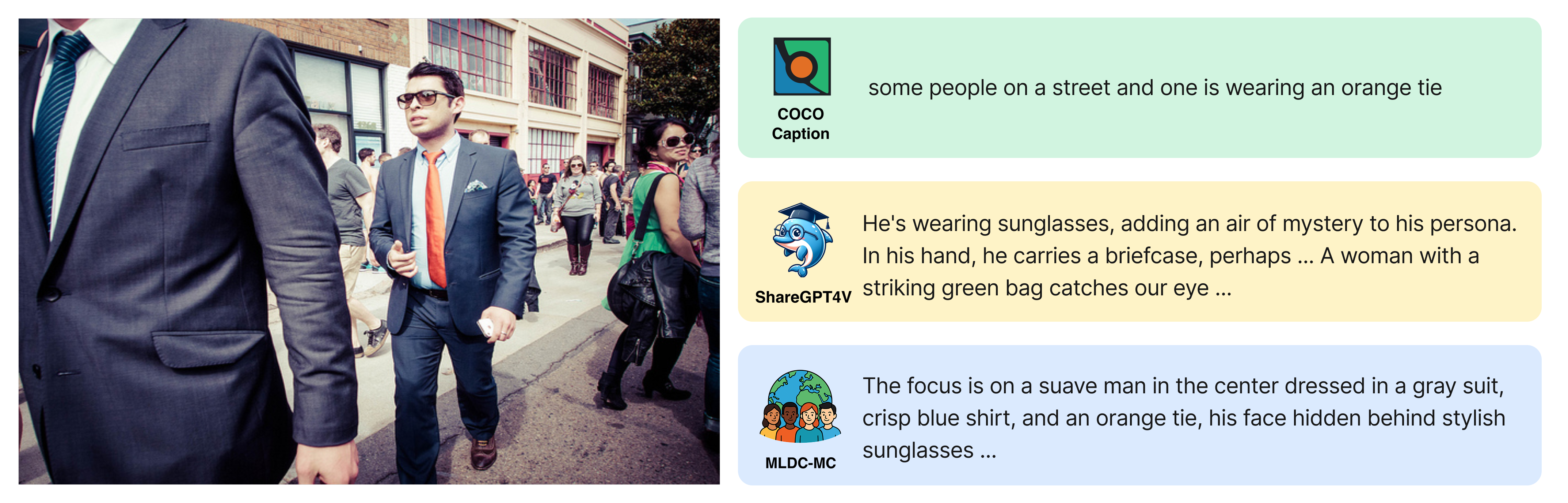}
  \caption{\textbf{Captions for an image from COCO, ShareGPT4V, and our MLDC-MC dataset.} The English annotation is summarized from three individual annotations and post-processed. The COCO caption lacks detail, while the ShareGPT4V caption contains several inaccuracies, which are in red. Our annotation is both detailed and accurate.}
  \label{fig:culture_part1}
\end{figure*}

\paragraph{Annotation Quality Assurance}
The platform enforces multiple quality control mechanisms to ensure annotation completeness and consistency. For 2D image tasks, the platform enforces that annotators mark at least five objects with native-language labels. For 3D scene tasks, the system implements a mandatory sequential workflow: individual object annotations (annotation step 1) must be completed first, then the system unlocks the scene-level recording interface (annotation step 2). This approach not only enables annotators to become familiar with the objects in the scene first, but also increases the density of the overall description. Temporal constraints enforce minimum recording durations (20 seconds for individual objects, 60 seconds for scenes or images) with automatic 3-minute cutoffs to balance annotation depth and annotator fatigue. Also, we developed a dual-transcript preservation mechanism that stores both the automatically generated transcription from OpenAI's Whisper API and the user-edited version, enabling us to detect excessive discrepancies between them and identify cases where annotators might have copied the VLMs' output instead of recording manually. At last, we use GPT-4o to summarize multiple transcriptions into one final caption to improve the text quality of the transcriptions.

\subsection{Multilingual Dense Captioning for Multicultural Images (MLDC-MC)}

\paragraph{Part A: Captions-Only Dataset} We aimed to select
images containing culturally distinctive elements for human annotation. Approximately 20\% of the images were sourced from MMID \citep{hewitt-etal-2018-learning}, 20\% from MS COCO \citep{chen2015microsoftcococaptionsdata}, and about 60\% from CVQA \citep{romero2024cvqaculturallydiversemultilingualvisual}; further details are provided in the Appendix.
We collected 6,220 distinct captions in 13 different languages. \Cref{tab:culture_part1} contains detailed statistics for each language. Our captions are significantly longer and more detailed than COCO captions; on average, our English captions are 820 characters long, compared to 52 for COCO. They are also human-annotated and contain fewer inaccuracies and hallucinations compared to ShareGPT4V. \Cref{fig:culture_part1} contains an example of this with an image from the COCO dataset. The COCO caption is very short and fails to capture the majority of the detail in the image, while the ShareGPT4V caption is more detailed but contains major inaccuracies. For instance, it refers to a woman with a green bag, but the woman in the background does not have a green bag. It also claims that the person in the foreground is carrying a briefcase, which is also untrue. Our human-annotated caption is able to capture a large amount of the detail in the picture while also not containing factual inaccuracies.

\begin{figure*}[!t]
  \includegraphics[width=\textwidth]{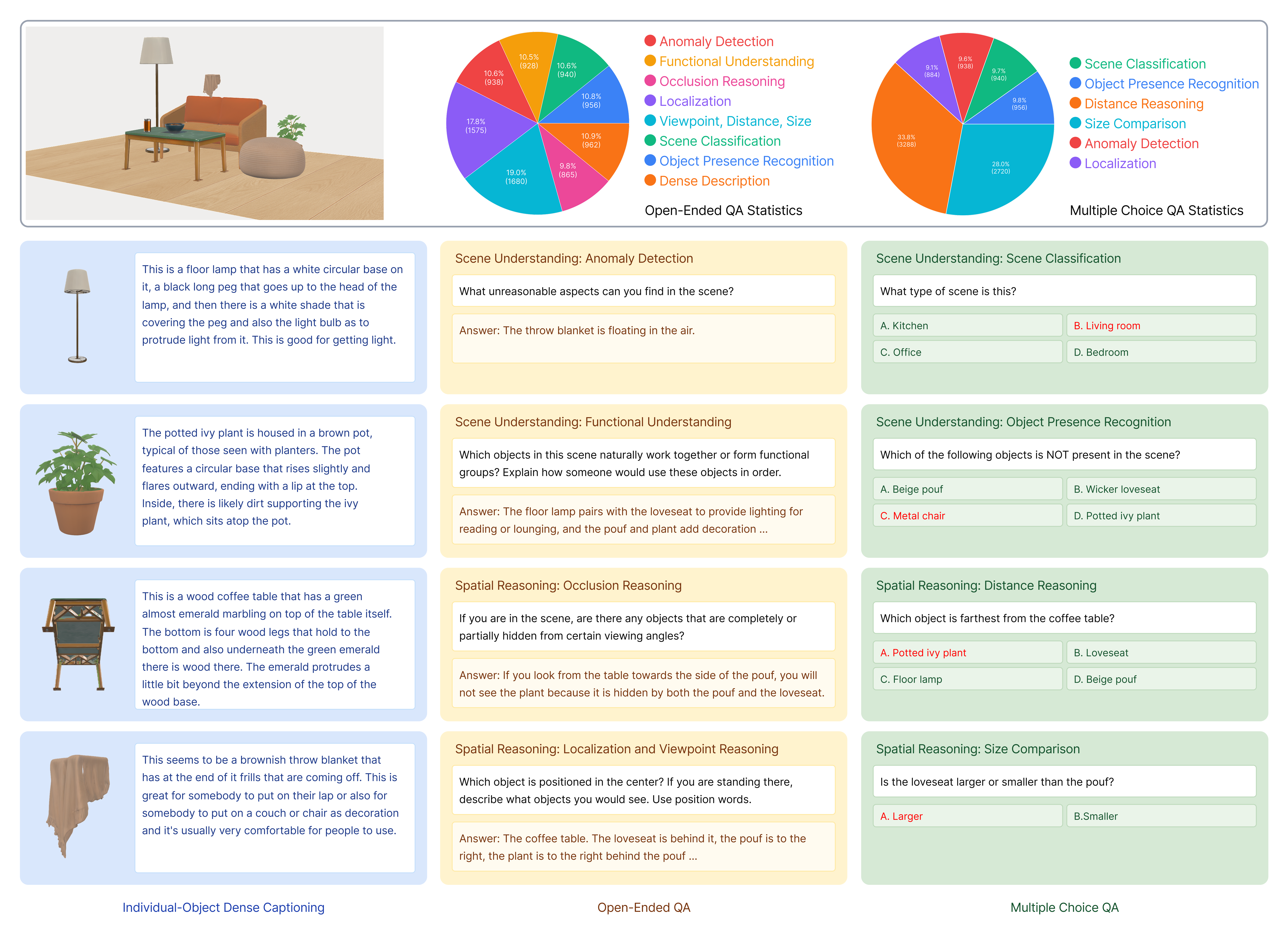}
  \caption{\textbf{Overview of the 3D scene dataset structure (MLDC-3D).} This figure selects one representative example to showcase the full scope of our dataset. The upper-left corner displays the 3D scene generated via HOLODECK 2.0. The data structure is presented in three columns: The first column illustrates Individual-Object Dense Captioning, providing fine-grained descriptions. The second column presents Open-Ended QA (OEQA) pairs derived from the dense transcriptions of human annotations. The third column shows the Multiple Choice QA (MCQA) generated based on the OEQA pairs. Note that the OEQA and MCQA examples shown in the figure only represent a subset of the available question types. The full set of categories and their distribution, which highlights the dataset's comprehensive coverage across various tasks (including scene understanding and spatial reasoning), is presented in the accompanying pie charts.}
  \label{fig:3D_1}
\end{figure*}

\paragraph{Part B: Captions-and-Points Dataset} In this part, our dataset is from Flickr. We collected 2,526 distinct captions in 18 different languages. To ensure diversity and cultural representation, we targeted images from India, Spain, and Korea. The search process relied on country-specific keywords (e.g., “street market India,” “cathedral Italy,” “festival Mexico”) to capture a wide range of cultural and everyday settings. Duplicates and low-quality images were removed, and only clear, contextually relevant samples were retained.

Images have been captioned by both native and non-native speakers. We hypothesize that dense captions created by native speakers include a lot of cultural information that would be missing in non-native captions. The pointing coordinates on the image are recorded as percentage values, retained to two decimal places.

\subsection{Multilingual Dense Captioning for 3D Scenes (MLDC-3D)}

We apply HOLODECK 2.0 \citep{bian2025holodeck20visionlanguageguided3d} to generate the 3D scenes. Then, we convert 3D scene models from GLB format to colored point clouds for further model training. All points are transformed to world coordinates, and the output is stored as an $N \times 6$ array (XYZ + RGB), where $N = 8{,}192$ points per scene.

To make the dataset diverse and robust, our design encompasses two primary types: indoor and outdoor, spanning across seven major categories (Home, Work Space, Commercial Space, Public Space, Nature, Urban Space, and Rural) and including 50 distinct scenes subcategories. Please refer to \cref{tab:scene-list} for details. In total, we collected the captions for 898 different 3D scenes and 7,460 different 3D objects, resulting in about 2,000 scene-level dense captions and 19,000 object-level dense captions. \Cref{tab:3d_scene_descriptions} contains detailed statistics for each language. For each of the scenes, we ask annotators to describe objects one by one in detail, and then answer 9 questions according to the scene. \Cref{tab:annotation-prompts-3D} shows the prompts for annotators.

Based on the original dense captions, we go further to generate open-ended and multiple-choice question-answer pairs. Stage 1 uses GPT-4o to extract structured answers from multi-language transcripts. By aggregating multiple transcripts in the same language for each scene, we enable GPT-4o to extract answers effectively (\cref{fig:3D_1}, Open-Ended QA). Specifically, all OEQA categories (e.g., Anomaly Detection) are sourced from the scene-level dense captions (from annotation step 2), with the exception of the 'Dense Description' category, which is derived from the individual-object captions (from annotation step 1). Because the answers are based on multiple transcripts, they are more comprehensive and reliable. Stage 2 converts OEQA pairs into multiple-choice questions using GPT-4o, leveraging question-specific strategies. For example, to ensure high-quality distractors, we sample them from cross-question data. This means that for options involving objects, we extract plausible distractors from the set of real objects identified in the scene via an Open-Ended QA ``Count and identify all the objects visible in the scene''. This yields 9,726 multiple-choice question-answer pairs across 6 variants.

\section{Multicultural and Multilingual Imagery Caption Generation}

In the following experiments, we fine-tune VLMs on our MLDC-MC dataset to demonstrate its effectiveness in improving multilingual generation, visual grounding, and multicultural alignment capabilities.

\subsection{Multilingual Generation Capability}
We use the pretrained Llama-3.2-11B-Vision-Instruct model as our base model and finetune it on our MLDC-MC Part A. We evaluate our finetuned model, MultilingualCap, on several captioning benchmarks to evaluate the quality of multilingual generation. We use XM3600 \citep{thapliyal2022crossmodal3600massivelymultilingualmultimodal} and xFlickrCO \citep{bugliarello2022igluebenchmarktransferlearning}, which are both multilingual image captioning datasets. XM3600 contains human-annotated
captions in 36 different languages and xFlickrCO has a mix of human-annotated and machine-translated captions in 8 different languages. We use standard textual similarity metric BERTScore \citep{zhang2020bertscoreevaluatingtextgeneration}. \Cref{tab:multilingual_comparison} reports the results. We also report the chrF++ \citep{popovic-2015-chrf} results broken down by language in \cref{tab:multilingual_lang_performance}.

Across both datasets, almost all of our metrics, and almost all of the languages we benchmark, MultilingualCap outperforms the base Llama model by a significant margin. This is an indication that models trained on our dataset are capable of producing superior multilingual captions. This is true even for languages that are not in our training dataset such as Indonesian, although the improvement is less significant. We hypothesize that our model learns a general, underlying structure for generating multilingual, detailed and culturally relevant captions, enabling effective cross-lingual transfer to unseen languages. On the other hand, our model demonstrates the largest performance gains in languages where the available training data volume is highest: MultilingualCap improves by 482\% over the baseline in Chinese and 1938\% in Russian in ChrF++ score on the xFlickrCO dataset.

\begin{table}[t]
\small
\centering
\begin{tabular}{llc}
\toprule
\textbf{Dataset} & \textbf{Model} & \textbf{BERTScore F$_1$} \\
\midrule
xFlickrCO & Llama-3.2-11B  & 0.777 \\
 & MultilingualCap & \textbf{0.819} \\
\midrule
XM3600 & Llama-3.2-11B  & 0.792 \\
 & MultilingualCap  & \textbf{0.822} \\
\bottomrule
\end{tabular}
\caption{\label{tab:multilingual_comparison}\textbf{Captioning evaluating results for our base model and multilingual variant, aggregated across all languages.}}
\end{table}

\subsection{Fine-Grained Visual Grounding Capability}
\label{sec:4.2}
We employ a two-stage fine-tuning strategy to adapt the Qwen2-VL-7B-Instruct vision-language model for multilingual multicultural dense captioning tasks.

\textbf{Stage 1: Multilingual Image-Text Alignment.} In the first stage, we pre-train the model on MLDC-MC Part A to establish robust cross-modal alignment between visual inputs and textual descriptions across multiple languages. This stage enables the model to enhance foundational multilingual vision-language understanding capabilities.

\textbf{Stage 2: Visual Grounding by Pointing.} The second stage fine-tunes the Stage 1 checkpoint on MLDC-MC Part B with point-level annotations. This stage focuses on refining the model's ability to generate precise, localized descriptions that correspond to specific regions of interest in the input images. The sequential training strategy allows the model to first acquire general multilingual capabilities before specializing in the target dense captioning task.

The point coordinates are extracted from MLDC-MC Part B where each marker point is represented as a tuple $(n, x, y)$, where $n$ denotes the object name and $(x, y)$ are normalized coordinates in percentage space $[0, 100]^2$. 



The assistant response is constructed by directly concatenating the descriptive transcript with the formatted point annotations, \eg, \texttt{a table with food\textbackslash n<point>65.20,63.90</point> table; <point>52.60,58.60</point> food;}. The human instruction explicitly prompts the model to follow this format: ``list keypoints as \texttt{<point>x,y</point> name} (percent coordinates) for all objects you are describing,'' enabling the model to learn this structured output format through supervised fine-tuning.

This text-based encoding approach enables the vision-language model to naturally learn the association between visual regions and their corresponding spatial coordinates through supervised fine-tuning, without requiring architectural modifications to handle coordinate outputs separately. 

We recruited five students to conduct a human evaluation along three dimensions:
\begin{itemize}
    \item \textbf{Point–Caption Consistency:} evaluates how well the point names align with the entities mentioned in the caption, reflecting semantic correspondence.
    \item \textbf{Spatial Accuracy:} measures how precisely the annotated points are localized on the objects mentioned in the point names.
    \item \textbf{Object Coverage Completeness:} assesses whether the annotated points adequately cover the objects in the images.
\end{itemize}

\begin{figure}[!t]
  \includegraphics[width=\columnwidth]{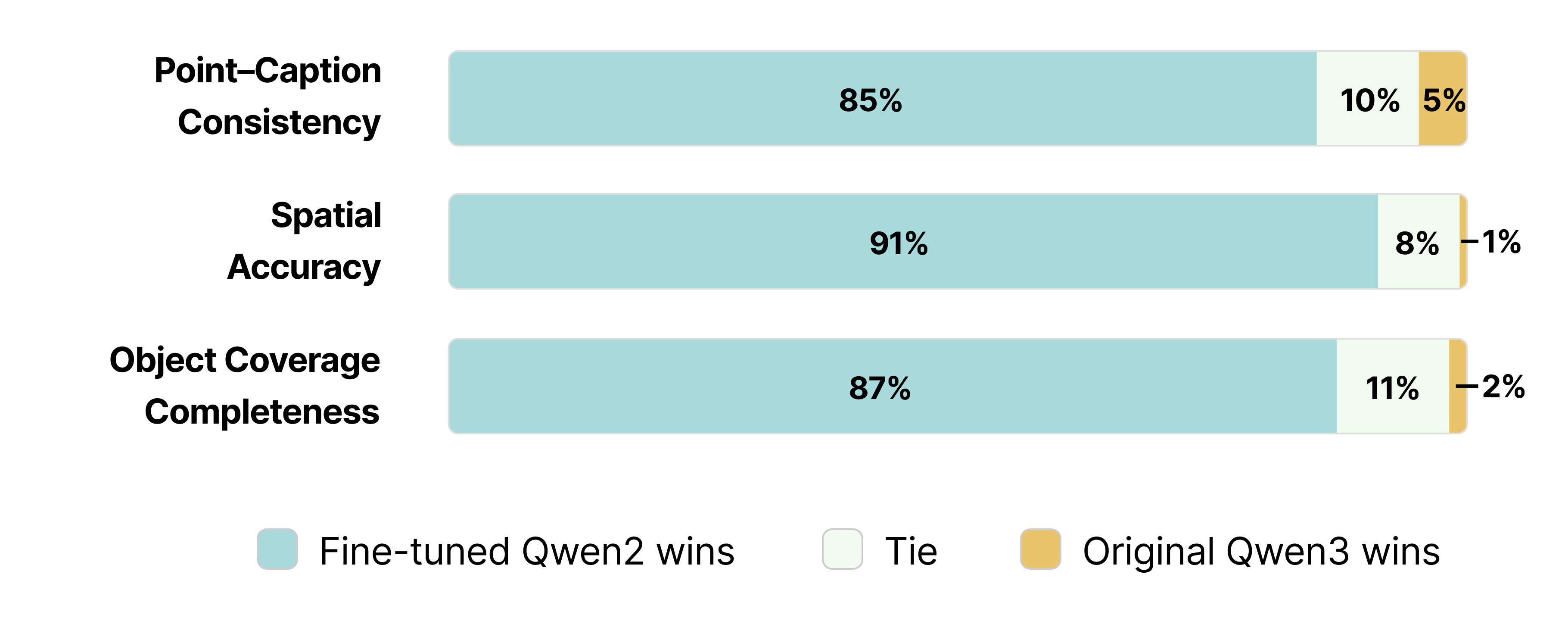}
  \caption{\textbf{Human evaluation results of visual grounding capability.} We compare the Qwen2 model fine-tuned on MLDC-MC with the vanilla Qwen3 on three dimensions: Point–Caption Consistency, Spatial Accuracy, and Object Coverage Completeness. Each bar shows the percentage of test set images.}
  \label{fig:pointing_compare}
\end{figure}

\begin{figure}[!t]
  \includegraphics[width=\columnwidth]{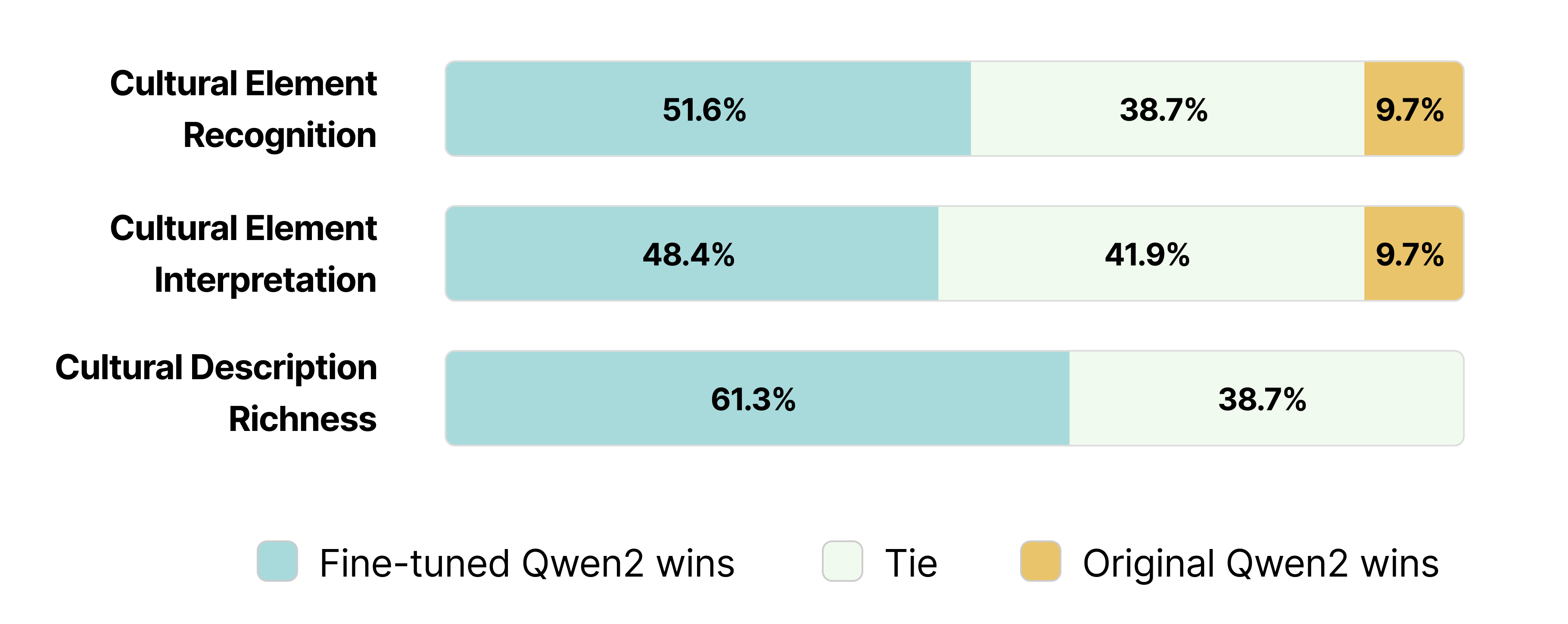}
  \caption{\textbf{Human evaluation results of cultural alignment capability.} We compare the Qwen2 model fine-tuned on Chinese data from MLDC-MC part A with the vanilla version on three dimensions: Cultural Element Recognition, Cultural Element Interpretation, and Cultural Description Richness. Each bar shows the percentage of test set images.}
  \label{fig:chinese}
\end{figure}

For each sample, evaluators are presented with the image and the outputs from the model before and after training. They are asked to determine, for each evaluation dimension, which version performs better or if they are equivalent. The final result for each image is determined by taking the majority vote among the evaluators. \Cref{fig:pointing_compare,fig:pointing} compares the test set outputs of Qwen2-VL-7B-Instruct (after two-stage training) and vanilla Qwen3-VL-8B-Instruct. It is evident that the fine-tuned Qwen2 outperforms the more advanced Qwen3. We also evaluated the vanilla Qwen2-VL-7B-Instruct model on this pointing task using the same prompts; however, it was unable to follow the keypoint-listing instruction or produce any coordinate outputs.

\subsection{Multicultural Alignment Capability}
\label{sec:4.3}
To further demonstrate that our dataset can enhance the multicultural capability of VLMs, we first examine its effectiveness on aligning with Chinese cultural contexts. 

We fine-tuned Qwen2-VL-7B-Instruct using only Chinese image-text pairs, which Chinese people label. Our MLDC-MC part A dataset includes a total of 631 Chinese-culture-related images annotated in Chinese, with 600 images used for training and 31 for testing. We compared the outputs of the Qwen model before and after training. To assess the cultural alignment quality, we recruited three native Chinese speakers to conduct a human evaluation along three dimensions:
\begin{itemize}
    \item \textbf{Cultural Element Recognition:} whether the model correctly identifies Chinese cultural elements present in the image.
    \item \textbf{Cultural Element Interpretation:} whether the model provides accurate and meaningful explanations of the identified cultural elements.
    \item \textbf{Cultural Description Richness:} whether the generated captions describe cultural elements in a detailed, diverse, and contextually informative manner.
\end{itemize}

The evaluation process is similar to the one in \cref{sec:4.2}. \Cref{fig:chinese} shows the human evaluation results comparing the fine-tuned Qwen model with the original version. The fine-tuned model consistently outperforms the baseline across all three dimensions. Specifically, it achieves higher accuracy in recognizing Chinese cultural elements, indicating that the model has learned to better identify culturally significant visual cues. The improvement in cultural element interpretation further suggests that the model not only detects these elements but also understands their underlying cultural meanings. Moreover, the fine-tuned model demonstrates a substantial advantage in cultural description richness, showing its ability to generate more detailed and contextually enriched cultural descriptions. Overall, these results verify that fine-tuning on MLDC-MC effectively enhances the model’s cultural alignment and descriptive capability in Chinese cultural contexts, and further indicate that the proposed dataset has the potential to enhance the multicultural capability of VLMs.

\section{Scene-level PointLLM}

To evaluate the impact of our dataset on 3D scene understanding and reasoning, we conduct experiments using the PointLLM framework. While prior work has primarily focused on single-object point clouds, our MLDC-3D dataset introduces rich scene-level supervision across multiple question types. These experiments further validate that MLDC-3D complements existing 3D instruction-tuning resources and enables robust learning of global scene semantics and complex spatial relationships.

\subsection{Model and Data Preprocessing}
A distinct advantage of our data is the inclusion of complete 3D scenes, unlike the 2D images or videos commonly derived from them. Therefore, we chose PointLLM for its capability to directly process 3D objects. PointLLM proposes a novel two-stage training process, leveraging a large, automatically generated dataset of point-text instructions, enabling the model to accurately classify and caption 3D objects.

To ensure compatibility with the PointLLM architecture, we maintain the same point cloud representation format as the original Objaverse-based dataset. Each scene is represented as an $N \times 6$ matrix stored in NumPy format (.npy). The key difference from the original PointLLM dataset lies in the semantic scope: while PointLLM uses single-object point clouds from Objaverse, our dataset contains multi-object scene-level point clouds where spatial relationships between objects are preserved through world coordinate transformations. 

Our scene-level instruction-following data adopts the Stage 2 conversation format from PointLLM. The dataset comprises two conversation types: \textit{detailed\_description} for comprehensive scene descriptions and \textit{single\_round} for question-answering tasks (including both open-ended questions and multiple-choice questions). Each sample follows the standard format with fields \texttt{object\_id} (scene name), \texttt{conversation\_type}, and \texttt{conversations} containing human-assistant dialogue pairs. The special token \texttt{<point>} is prepended to the first user query to indicate point cloud input. This format ensures seamless integration with PointLLM's data loader.

We partition the MLDC-3D dataset to ensure no data leakage between training and testing. Our dataset comprises 898 unique scenes, which are split into 798 training scenes and 100 test scenes, following a scene-balanced strategy. Specifically, we ensured that two scenes from each of the 50 scene subcategories were reserved for the test set. This results in 16,485 training samples (combining 7,854 open-ended QA and 8,631 multiple-choice QA) and 2,085 test samples (combining 990 open-ended QA and 1,095 multiple-choice QA), maintaining an approximate 8:1 train-test ratio.

\subsection{Experimental Results and Analysis}

\textbf{Baseline Performance}
To establish a baseline, we first evaluate $\text{PointLLM\_7B\_v1.2}$ \citep{xu2024pointllmempoweringlargelanguage} using our multiple-choice QA. However, qualitative analysis of the generated outputs reveals a critical insight: the original model produces degenerate outputs consisting of repetitive tokens, special characters, and incoherent text fragments (\eg, \texttt{Sent\S\#H\~{}thexic}). In other words, the baseline model exhibits \textbf{complete generative failure} on our scene tasks. This finding suggests that the pre-trained model, despite being trained on 660K brief description data and 70K complex instruction data, lacks the training distribution alignment necessary for our scene-level tasks. This inherent limitation underscores the importance of our novel dataset, which is specifically designed to drive advancements in scene-level 3D understanding and complex reasoning. We train the model with one sixty-fourth of the training data to provide basic scene capacity and use it as the baseline.

\begin{figure}[!t]
  \includegraphics[width=\columnwidth]{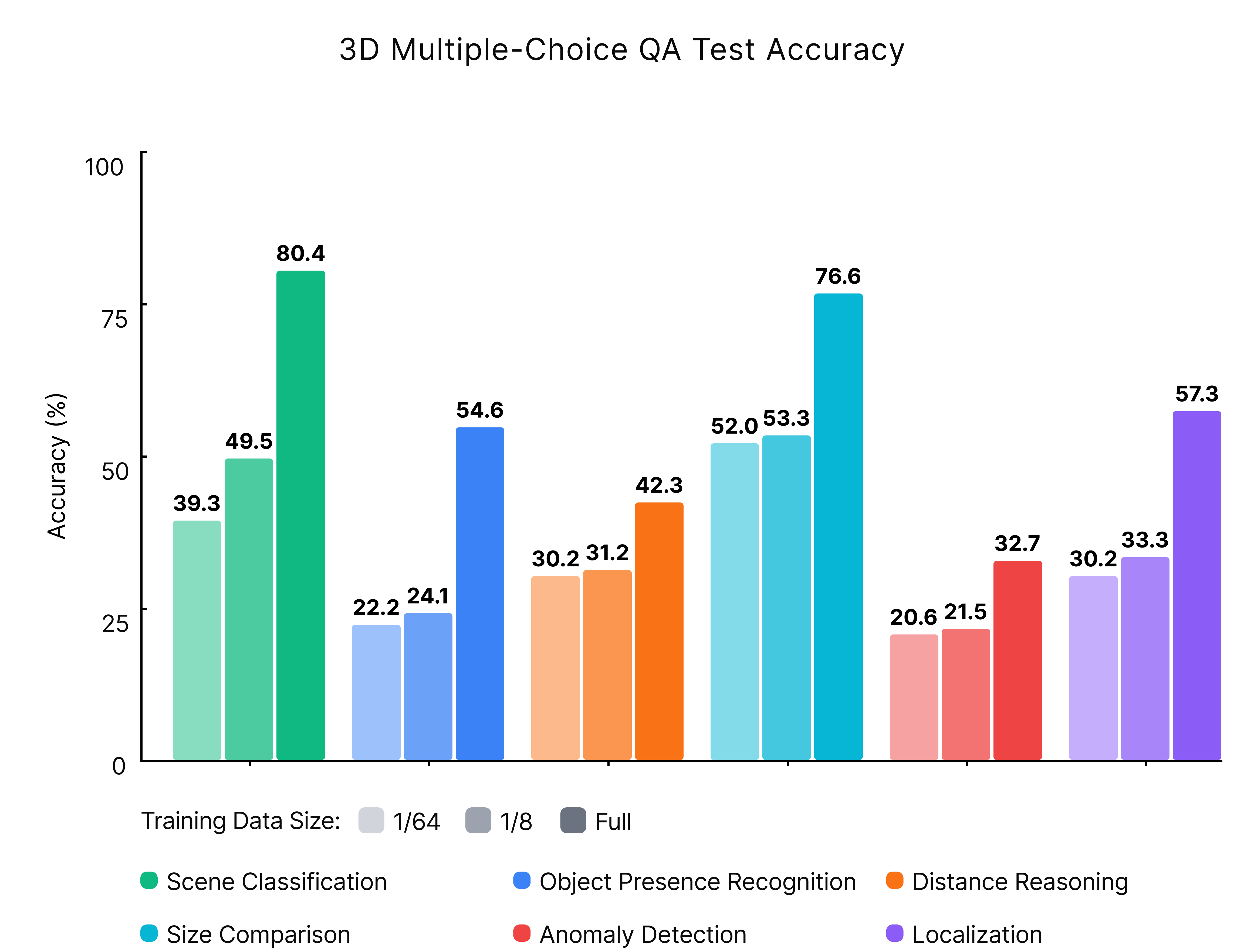}
  \caption{\textbf{Performance breakdown by question category on the MLDC-3D multiple-choice QA test set.} This figure presents the accuracy of the fine-tuned PointLLM model across the six distinct question categories of the MLDC-3D dataset. Clearly, performance improves substantially with more training data.}
  \label{fig:3Dtest}
\end{figure}

\textbf{Scene-level PointLLM} We finetune $\text{PointLLM\_7B\_v1.2}$ using 16,485 training samples. Our fine-tuned model generates clean, concise, and well-formatted responses and demonstrates a transformation from model collapse to functional task execution. The multiple-choice QA test set performance is shown in \cref{fig:3Dtest}. Our fully fine-tuned model achieves an overall accuracy of 57.26\%, substantially outperforming the baseline of 37.17\%. Performance varies significantly across question types, revealing distinct strengths and weaknesses of the learned representations.

The model demonstrates strong performance on two tasks: scene classification achieves 80.37\% accuracy, and size comparison reaches 76.62\%, indicating effective learning of global scene semantics and spatial scale relationships. Medium-difficulty tasks, including localization (57.29\%) and object presence recognition (54.63\%), show reasonable but improvable performance. Notably, the model struggles with tasks requiring complex spatial understanding or reasoning: distance reasoning achieves only 42.28\% accuracy, while anomaly detection, which requires common-sense reasoning and scene context understanding, proves most challenging at 32.71\%, only marginally above random chance.

The performance disparity across categories suggests that: (1)~the model successfully captures coarse-grained scene-level features but requires further refinement for fine-grained spatial relationship understanding and reasoning, and (2)~incorporating external knowledge or multi-modal reasoning may be necessary for tasks demanding common-sense inference. The largest category (distance relations, 3,288 training samples), showing suboptimal performance, indicates a critical area for model improvement. These results establish a strong baseline for 3D scene understanding and reasoning using point cloud-based LLMs, and highlight promising directions for future architectural and training enhancements.

\section{Conclusion}
We introduced DenseAnnotate, an audio-driven multimodal annotation platform that scales dense caption collection across both 2D imagery and 3D scenes. Our MLDC-MC and MLDC-3D datasets address critical gaps in multilingual, multicultural, and 3D scene data resources. Empirical evaluations demonstrate that models fine-tuned on these datasets achieve substantial gains. By integrating human expressiveness with scalable automation, DenseAnnotate offers a practical and extensible paradigm for next-generation vision–language research, setting a foundation for more inclusive multimodal intelligence.
{
    \small
    \bibliographystyle{ieeenat_fullname}
    \bibliography{main}
}
\clearpage
\setcounter{page}{1}
\maketitlesupplementary

\section{DenseAnnotate Platform}
\label{sec:DenseAnnotate Platform}
\begin{figure*}[h]
  \includegraphics[width=\textwidth]{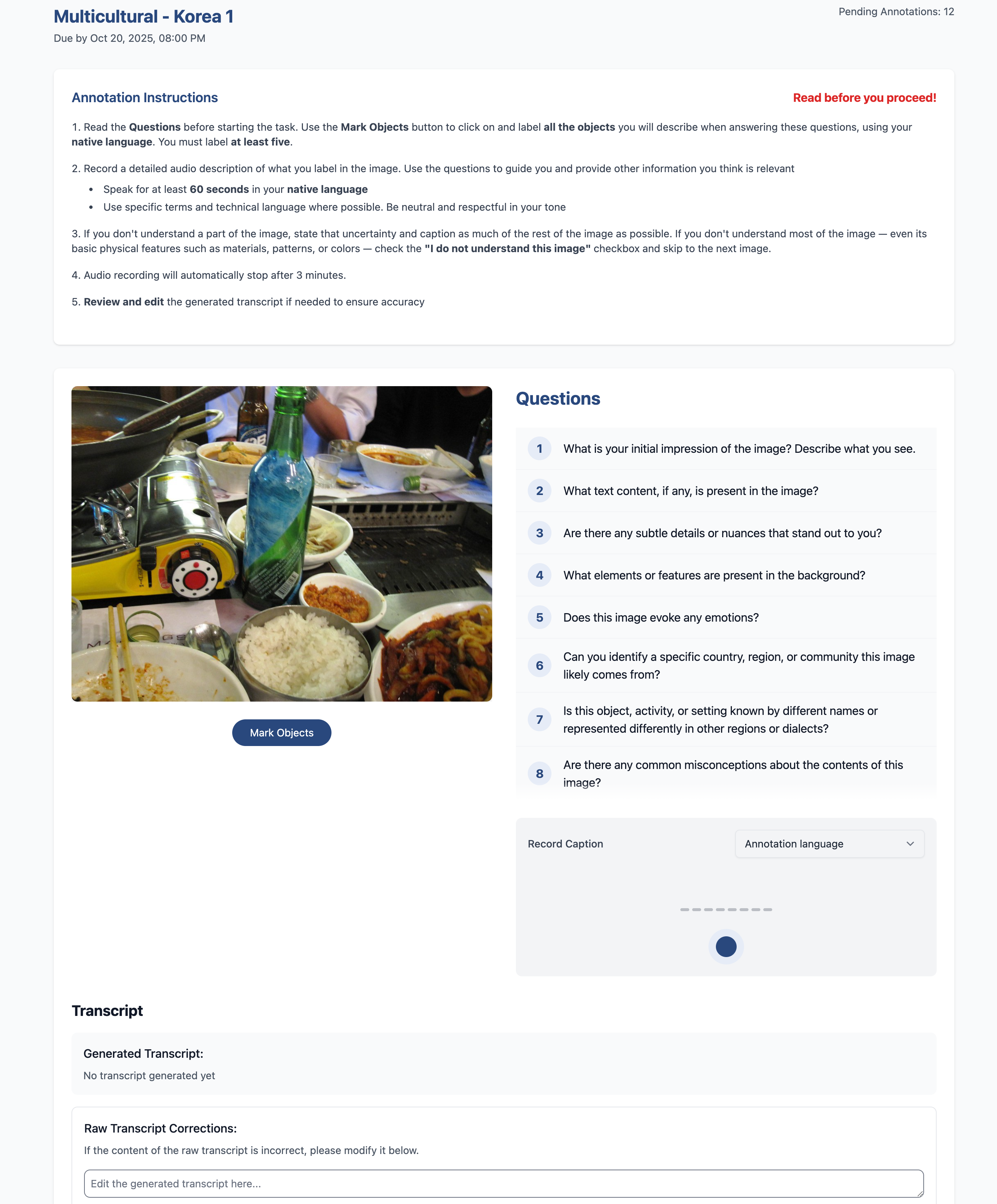}
  \caption{Screenshot of the DenseAnnotate platform showing the image dense captioning module.}
  \label{fig:platform_cultural}
\end{figure*}

\begin{figure*}[h]
  \includegraphics[width=\textwidth]{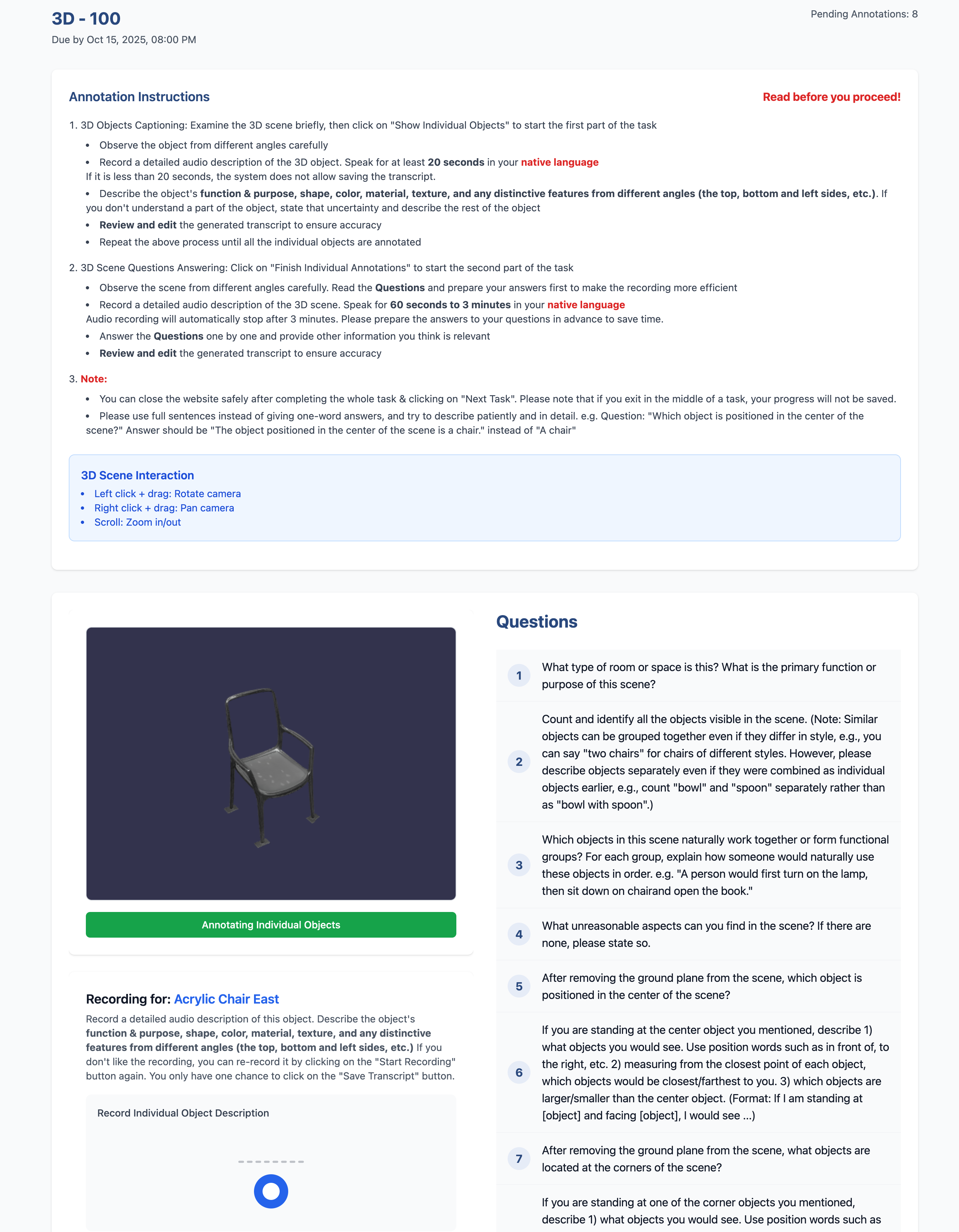}
  \caption{Screenshot of the DenseAnnotate platform showing the 3D dense captioning module.}
  \label{fig:platform_3d}
\end{figure*}

Screenshots of our platform are displayed in \cref{fig:platform_cultural,fig:platform_3d}.

\paragraph{System Architecture and Design}
We developed a web-based multimodal annotation platform that unifies data collection workflows for both 2D images and interactive 3D scenes. We collect audio data rather than textual data from annotators, which has been shown to improve the quality and ease of collection of dense captions \citep{deitke2024molmopixmoopenweights}. The system employs a three-tier architecture consisting of a React-based frontend for real-time annotation interfaces, a Flask backend integrated with Supabase Backend-as-a-Service for scalable data management, and a PostgreSQL relational database with row-level security policies ensuring multi-tenant data isolation. Our design prioritizes organizational scalability through a hierarchical model where administrators create tasks with customizable questions and instructions, which are then distributed to annotators based on demographic metadata and organizational membership. The architecture incorporates Babylon.js WebGL rendering engine for client-side 3D visualization, enabling annotators to interact with 3D scenes through intuitive camera controls (rotation, panning, zoom) without requiring specialized software installation.

\section{MLDC-MC Data}
\begin{figure*}[t]
  \includegraphics[width=\textwidth]{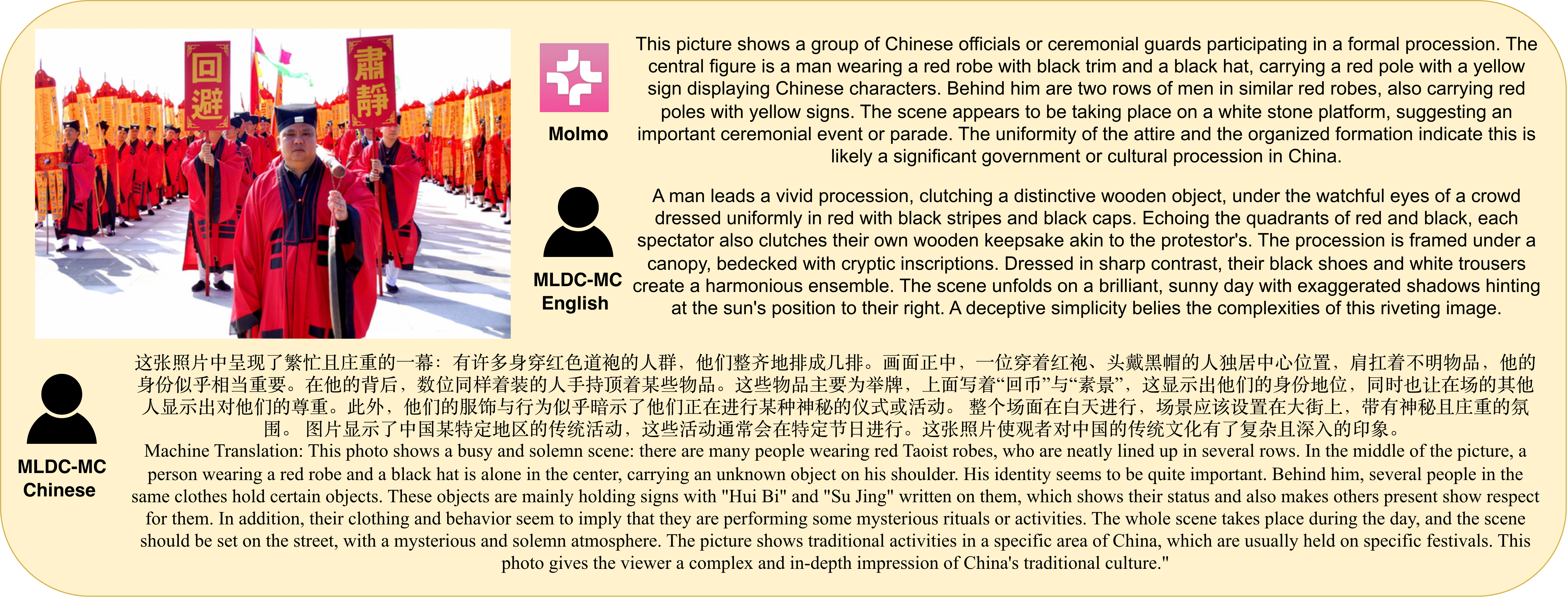}
  \caption{Caption outputs on a culturally distinct image for Molmo, an English annotation from our dataset, and a Chinese annotation from our dataset. Molmo was prompted with "What is this image?". The English and Chinese annotations are summarized from three individual annotations and post-processed. The English translation of the Chinese annotation is also depicted. While Molmo and the English annotator fail to capture the cultural nuance in the photo, our native Chinese-speaking annotator does. This is an illustrative example of the importance of using native speakers for culturally distinctive images.}
  \label{fig:culture_part2}
\end{figure*}

\begin{figure*}[t]
  \includegraphics[width=\textwidth]{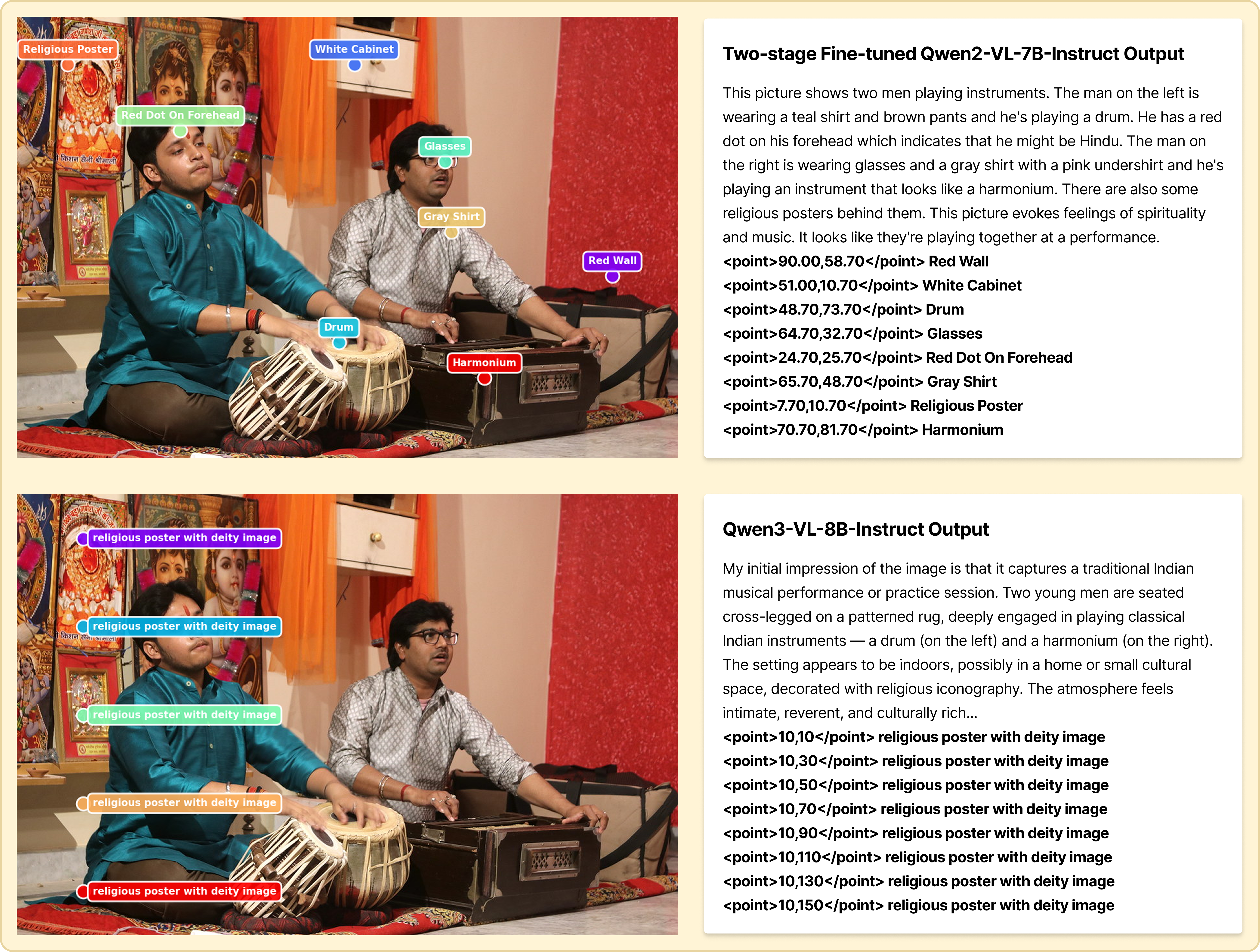}
  \caption{Comparison of test outputs from fine-tuned Qwen2 and vanilla Qwen3. The Qwen2 fine-tuned on our data demonstrates the ability to identify cultural information, generate dense captions, and accurately localize points within the image. Using the same prompt, Qwen3 produces repeated points and outputs values exceeding the maximum percentage coordinate (100\%), resulting in a poor alignment with the caption output.}
  \label{fig:pointing}
\end{figure*}

The statistics of the languages in MLDC-MC are shown in \cref{tab:culture_part1,tab:stage2_all_languages}. The annotation prompts are shown in \cref{tab:annotation-prompts}. \Cref{fig:culture_part2} gives an example of our  MLDC-MC data, which shows our data captures the cultural information. The test result of multilingual generation capability is shown in \cref{tab:multilingual_lang_performance}. \Cref{fig:pointing} demonstrates the performance of pointing and captioning after fine-tuning on our data.

\section{MLDC-3D Data}
The statistics of the languages in MLDC-3D are shown in \cref{tab:3d_scene_descriptions}. The annotation prompts are shown in \cref{tab:annotation-prompts-3D}. \Cref{tab:scene-list} shows all the 3D scenes types in the data.

\paragraph{Data Acquisition}

We apply HOLODECK 2.0 \citep{bian2025holodeck20visionlanguageguided3d} to generate the 3D scenes. The process begins with a natural language description which is processed by a Vision-Language Model (VLM) to generate a 2D reference image for style and then extract quality-controlled, individual 2D images for each object. These 2D images are fed into 3D generative models (such as Hunyuan3D 2.1) to efficiently create high-quality 3D assets. Finally, the VLM infers spatial constraints from the text and image, which are iteratively applied by a Depth-First-Search (DFS) solver to achieve a semantically coherent and physically plausible 3D layout.

We convert 3D scene models from GLB format to colored point clouds using Blender's Python API for further model training. After importing the GLB file and triangulating all mesh objects (excluding the ground plane), we perform surface sampling using barycentric coordinates. For each sampled point on a triangular face, we generate random barycentric weights ($r_1, r_2, r_3$) with the constraint that if $r_1 + r_2 > 1$, then $r_1 \leftarrow 1 - r_1, r_2 \leftarrow 1 - r_2$, and compute the position as $p = r_1v_1 + r_2v_2 + r_3v_3$. For color information, we interpolate UV coordinates using the same barycentric weights and perform nearest-neighbor sampling from the texture maps. All points are transformed to world coordinates, and the output is stored as an $N \times 6$ array (XYZ + RGB), where $N = 8{,}192$ points per scene.

\section{Ethics Statement}
This study involves data collected from student participants at our institution. The data collection was conducted in accordance with our institution’s ethical guidelines but was not subject to a formal Institutional Review Board (IRB) approval process. All participants were informed about the purpose of the study and provided their consent for their data to be used for research purposes. The collected data contain no personally identifiable or sensitive information, and participation was entirely voluntary.

\FloatBarrier

\begin{table*}
  \small
  \centering
  \begin{tabular}{lccccccccccccc}
    \toprule
    \textbf{} & \textbf{zh} & \textbf{en} & \textbf{ko} & \textbf{ru} & \textbf{hi} & \textbf{ta} & \textbf{ja} & \textbf{vi} & \textbf{es} & \textbf{no} & \textbf{de} & \textbf{th} & \textbf{ur} \\
    \midrule
    \textbf{Annotation Count}      & 2777 & 2427 & 378 & 149 & 122 & 107 & 75  & 65  & 51  & 26  & 21  & 12  & 10 \\
    \textbf{Median Word Count}     & 1$^{\dagger}$ & 157 & 104 & 110 & 169 & 85  & 1$^{\dagger}$ & 134 & 97  & 135 & 122 & 1$^{\dagger}$  & 72 \\
    \textbf{Median Character Count}& 222  & 820  & 410 & 677 & 741 & 639 & 309 & 603 & 517 & 717 & 754 & 832 & 318 \\
    \bottomrule
  \end{tabular}
  \caption{\label{tab:culture_part1}
    Number of annotations and median length for our collected captions across all languages in the MLDC-MC part A. $^{\dagger}$For Chinese, Japanese, and Thai, we report a median word count of 1 because these languages do not use whitespace to delimit words; the more informative statistic is the character count in the third row. Please refer to the Appendix for the statistics of MLDC-MC Part B.
  }
\end{table*}

\begin{table*}
  \centering
  \begin{tabular}{lcccccccccc}
    \hline
    \textbf{} & \textbf{en} & \textbf{zh} & \textbf{ko} & \textbf{vi} & \textbf{es} & \textbf{hi} & \textbf{he} & \textbf{ja} & \textbf{ru} & \textbf{ne} \\
    \hline
    \textbf{Annotation Count}      & 1047 & 971 & 148 & 45  & 43  & 42  & 40  & 32  & 29  & 26 \\
    \textbf{Median Word Count}     & 159  & 1$^{\dagger}$ & 77  & 136 & 134 & 145 & 129 & 1$^{\dagger}$ & 95  & 93 \\
    \textbf{Median Character Count}& 684  & 222  & 226 & 444 & 577 & 568 & 577 & 288 & 575 & 415 \\
    \hline
  \end{tabular}
  
  \vspace{0.3cm}
  
  \begin{tabular}{lcccccccc}
    \hline
    \textbf{} & \textbf{ur} & \textbf{ar} & \textbf{ro} & \textbf{gu} & \textbf{pt} & \textbf{te} & \textbf{ta} & \textbf{th} \\
    \hline
    \textbf{Annotation Count}      & 15  & 15  & 15  & 14  & 14  & 14  & 13  & 3 \\
    \textbf{Median Word Count}     & 166 & 73  & 81  & 133 & 129 & 45  & 110 & 1$^{\dagger}$ \\
    \textbf{Median Character Count}& 540 & 298 & 436 & 512 & 586 & 278 & 827 & 813 \\
    \hline
  \end{tabular}
  \caption{\label{tab:stage2_all_languages}
    Number of annotations and median length for our collected captions across all languages in the MLDC-MC part B. $^{\dagger}$For Chinese, Japanese, and Thai, we report a median word count of 1 because these languages do not use whitespace to delimit words; the more informative statistic is the character count in the third row.
  }
\end{table*}

\begin{table*}
  \centering
  \begin{tabular}{p{8cm} p{8cm}}
    \hline
    \textbf{MLDC-MC Part A} & \textbf{MLDC-MC Part B}\\
    \hline
    What is the image at first glance? &
    What is your initial impression of the image? Describe what you see. \\\hline

    What are the objects and their counts? &
    What text content, if any, is present in the image? \\\hline

    What does the text say? &
    Are there any subtle details or nuances that stand out to you?\\\hline

    What are the positions of the objects? &
    What elements or features are present in the background? \\\hline

    What subtle details are noticeable? &
    Does this image evoke any emotions? \\\hline

    What is in the background? &
    Can you identify a specific country, region, or community this image likely comes from? \\\hline

    What is the style and color? &
    Is this object, activity, or setting known by different names or represented differently in other regions or dialects? \\\hline

    Is there any contextual information such as location that might help to understand the image? &
    Are there any common misconceptions about the contents of this image?  \\\hline

    Is there anything that can be inferred from the image? &
    --- \\\hline

    Is this image culturally distinct? If yes, please explain why you think this image is culturally distinct. &
    --- \\
    \hline
  \end{tabular}
  \caption{\label{tab:annotation-prompts}
    Annotation prompts for MLDC-MC.
  }
\end{table*}

\begin{table*}[t]
\centering
\begin{tabular}{llcc}
\toprule
\textbf{Dataset} & \textbf{Lang} & \textbf{Llama-3.2-11B-Vision-Instruct} & \textbf{MultilingualCap} \\
\midrule
xFlickrCO & de & 5.911 & \textbf{7.219} \\
 & es & 9.531 & \textbf{12.069} \\
 & id & 8.616 & \textbf{10.133} \\
 & ja & 2.237 & \textbf{4.922} \\
 & ru & 0.645 & \textbf{13.144} \\
 & tr & 5.890 & \textbf{8.423} \\
 & zh & 0.382 & \textbf{2.225} \\
\cline{1-4}
XM3600 & ar & 1.080 & \textbf{4.683} \\
 & bn & 6.363 & \textbf{8.024} \\
 & cs & 7.553 & \textbf{8.900} \\
 & da & \textbf{8.958} & 8.432 \\
 & de & \textbf{9.187} & 7.378 \\
 & el & 3.691 & \textbf{3.846} \\
 & es & \textbf{9.514} & 7.690 \\
 & fa & 3.931 & \textbf{6.637} \\
 & fi & 5.157 & \textbf{7.955} \\
 & fil & 10.736 & \textbf{15.606} \\
 & fr & 8.370 & \textbf{9.187} \\
 & fuj & 1.764 & \textbf{2.782} \\
 & he & 5.745 & \textbf{7.264} \\
 & hi & 1.218 & \textbf{8.825} \\
 & hr & 7.115 & \textbf{8.183} \\
 & huo & 3.472 & \textbf{4.807} \\
 & id & 10.386 & \textbf{13.187} \\
 & it & 10.798 & \textbf{13.091} \\
 & ja & 3.034 & \textbf{3.314} \\
 & mi & 3.127 & \textbf{3.958} \\
 & nl & 4.195 & \textbf{6.105} \\
 & no & 5.885 & \textbf{6.411} \\
 & pl & 5.889 & \textbf{9.090} \\
 & pt & \textbf{9.719} & 9.556 \\
 & quz & \textbf{3.954} & 2.599 \\
 & ro & 8.048 & \textbf{8.177} \\
 & ru & 0.357 & \textbf{8.708} \\
 & sv & \textbf{5.817} & 5.239 \\
 & sw & \textbf{10.895} & 9.986 \\
 & te & \textbf{8.987} & 8.403 \\
 & th & 6.403 & \textbf{9.721} \\
 & tr & 4.649 & \textbf{5.387} \\
 & uk & 5.336 & \textbf{10.506} \\
 & vi & 5.942 & \textbf{13.542} \\
 & zh & 0.123 & \textbf{4.826} \\
\bottomrule
\end{tabular}
\caption{\label{tab:multilingual_lang_performance} chrF++ scores for each model across all languages in the two evaluation datasets. Note that chrF++ is not necessarily comparable across languages.}
\end{table*}

\FloatBarrier

\begin{table*}
  \centering
  \small
  \begin{tabular}{lcccccccccc}
    \toprule
    \textbf{} & \textbf{en} & \textbf{zh} & \textbf{hi} & \textbf{ko} & \textbf{vi} & \textbf{pt} & \textbf{id} & \textbf{ru} & \textbf{ur} & \textbf{es} \\
    \midrule
    \textbf{Annotation Count}      & 1004 & 736 & 32  & 27  & 25  & 24  & 16  & 16  & 16  & 12 \\
    \textbf{Median Word Count}     & 855  & 1$^{\dagger}$ & 661 & 449 & 785 & 670 & 776 & 383 & 758 & 506 \\
    \textbf{Median Character Count}& 3568 & 1210 & 2252 & 1385 & 2586 & 3011 & 4483 & 2189 & 2351 & 2266 \\
    \bottomrule
  \end{tabular}
  \caption{\label{tab:3d_scene_descriptions}
    Number of annotations and median length for our collected captions across the top 10 languages by sample count in the MLDC-3D. For each scene annotation, we concatenate the object-level and scene-level captions to calculate the length. The dataset contains 26 languages. $^{\dagger}$For Chinese, we report a median word count of 1 because these languages do not use whitespace to delimit words; the more informative statistic is the character count in the third row.
  }
\end{table*}

\begin{table*}[t]
  \centering
  \begin{tabular}{p{16cm}}
    \hline
    \textbf{MLDC-3D}\\
    \hline
     What type of room or space is this? What is the primary function or purpose of this scene?\\\hline

    Count and identify all the objects visible in the scene. (Note: Similar objects can be grouped together even if they differ in style, e.g., you can say "two chairs" for chairs of different styles. However, please describe objects separately even if they were combined as individual objects earlier, e.g., count "bowl" and "spoon" separately rather than as "bowl with spoon".)\\\hline

    Which objects in this scene naturally work together or form functional groups? For each group, explain how someone would naturally use these objects in order. e.g. "A person would first turn on the lamp, then sit down on chairand open the book."\\\hline

     What unreasonable aspects can you find in the scene? If there are none, please state so.\\\hline

    After removing the ground plane from the scene, which object is positioned in the center of the scene?\\\hline

    If you are standing at the center object you mentioned, describe 1) what objects you would see. Use position words such as in front of, to the right, etc. 2) measuring from the closest point of each object, which objects would be closest/farthest to you. 3) which objects are larger/smaller than the center object. (Format: If I am standing at [object] and facing [object], I would see ...)\\\hline

     After removing the ground plane from the scene, what objects are located at the corners of the scene?\\\hline

    If you are standing at one of the corner objects you mentioned, describe 1) what objects you would see. Use position words such as in front of, to the right, etc. 2) measuring from the closest point of each object, which objects would be closest/farthest to you. 3) which objects are larger/smaller than the center object. (Format: If I am standing at [object] and facing [object], I would see ...)\\\hline

     If you are in the scene, are there any objects that are completely or partially hidden from certain viewing angles? Describe the situation in detail. (Format: If I stand/sit/kneel/... at [object], facing [object], I can not see ..., because ...)\\
    \hline
  \end{tabular}
  \caption{\label{tab:annotation-prompts-3D}
    Annotation prompts MLDC-3D.
  }
\end{table*}

\begin{table*}
  \centering
  \begin{tabular}{llcc}
    \hline
    \textbf{Type} & \textbf{Category} & \multicolumn{2}{c}{\textbf{Subcategories}} \\
    \hline
    Indoor & Home & Bedroom & Living Room \\
           &      & Kitchen & Dining Room \\
           &      & Study Room & Bathroom \\
           &      & Children's Room & Balcony \\

    Indoor & Work Space & Office & Meeting Room \\
           &            & Library & Study Hall \\
           &            & Classroom & Laboratory \\
           &            & Recording Studio & Hospital Ward \\

    Indoor & Commercial Space & Coffee Shop & Restaurant \\
           &                  & Supermarket & Bar \\
           &                  & Bookstore & Indoor Market \\
           &                  & Hair Salon & Clinic \\

    Indoor & Public Space & Museum & Church \\
           &               & Theater & Music Room \\
           &               & Activity Room & Indoor Workshop \\
           &               & Indoor Flower Exhibition Hall & Gym \\
    \hline
    Outdoor & Nature & Beach & Forest Camping Site \\
            &       & Garden & Mountain Cabin \\
            &       & Desert Oasis & Lake Side \\

    Outdoor & Urban Space & Amusement Park & City Square \\
            &             & Bus Terminal & Rooftop Viewpoint \\
            &             & School Playground & Pedestrian Street \\

    Outdoor & Rural & Farmland & Ranch \\
            &       & Fishing Village Dock & Mountain Village \\
            &       & Marketplace & Orchard \\
    \hline
  \end{tabular}
  \caption{\label{tab:scene-list}
    The indoor and outdoor 3D scenes we use in this paper. For each scene, we generated ten instances using the Holodeck 2.0 model.
  }
\end{table*}


\end{document}